\documentclass[journal,twoside]{IEEEtran}
\ifCLASSINFOpdf
\else
\fi
\usepackage{verbatim}
\usepackage{amsmath}
\usepackage{amssymb}
\usepackage[slantedGreek]{mathptmx}
\usepackage{threeparttable}
\usepackage{cite}
\usepackage{enumerate}
\usepackage{color}
\usepackage{psfrag}
\usepackage{ifpdf}
\usepackage{multirow}
\usepackage{setspace}
\usepackage{url}
\usepackage{algorithmic}
\usepackage{algorithm}
\usepackage{booktabs}
\usepackage{array}
\usepackage{amsfonts}
\usepackage{graphicx}
\usepackage[numbers,sort&compress]{natbib}
\usepackage[caption=false,font=footnotesize]{subfig}
\usepackage{rotating}
\usepackage{textcomp}
\usepackage{tabularx}
\usepackage{bm}

\newcolumntype{L}[1]{>{\raggedright\arraybackslash}p{#1}}
\newcolumntype{C}[1]{>{\centering\arraybackslash}p{#1}}
\newcolumntype{R}[1]{>{\raggedleft\arraybackslash}p{#1}}

\begin{document}
%
\title{CRDN: Cascaded Residual Dense Networks for Dynamic MR Imaging with Edge-enhanced Loss Constraint}
%
%
%

\author{Ziwen Ke, Shanshan~Wang,~\IEEEmembership{Member,~IEEE,}
     Huitao Cheng, Qiegen Liu, Leslie Ying, Hairong Zheng 
    and~Dong Liang,~\IEEEmembership{Senior Member,~IEEE}
\thanks{This research was partly supported by the National Natural Science Foundation of China (61601450, 61871371, 81830056, 61771463, 61471350), Science and Technology Planning Project of Guangdong Province (2017B020227012), the Basic Research Program of Shenzhen (JCYJ20150831154213680), and Key Laboratory for Magnetic Resonance and Multimodality Imaging
of Guangdong Province (2017YFC0108802). * indicates the corresponding author.}
\thanks{Ziwen Ke, Shanshan~Wang contributed equally to this manuscript}
\thanks{Z. Ke, S. Wang, H. Cheng, H. Zheng and D. Liang are with Paul C. Lauterbur Research Center for Biomedical Imaging, Shenzhen Institutes of Advanced Technology, Chinese Academy of Sciences, Shenzhen 518055, China (e-mail: zw.ke@siat.ac.cn; sophiasswang@hotmail.com; dong.liang@siat.ac.cn).}
\thanks{Q. Liu is with the Department of Electronic Information Engineering, Nanchang University, Nanchang 330031, China.} 
\thanks{L. Ying is with the Department of Biomedical Engineering and the Department of Electrical Engineering, University at Buffalo, The State University of New York, Buffalo, NY 14260 USA.}

}

\maketitle

\begin{abstract}
 Dynamic magnetic resonance (MR) imaging has generated great research interest, as it can provide both spatial and temporal information for clinical diagnosis. However, slow imaging speed or long scanning time is still one of the challenges for dynamic MR imaging. Most existing methods reconstruct Dynamic MR images from incomplete k-space data under the guidance of compressed sensing (CS) or low rank theory, which suffer from long iterative reconstruction time. Recently, deep learning has shown great potential in accelerating dynamic MR. Our previous work proposed a dynamic MR imaging method with both k-space and spatial prior knowledge integrated via multi-supervised network training. Nevertheless, there was still a certain degree of smooth in the reconstructed images at high acceleration factors. In this work, we propose cascaded residual dense networks for dynamic MR imaging with edge-enhance loss constraint, dubbed as CRDN. Specifically, the cascaded residual dense networks fully exploit the hierarchical features from all the convolutional layers with both local and global feature fusion. We further utilize the total variation (TV) loss function, which has the edge enhancement properties, for training the networks. 
\end{abstract}

\begin{IEEEkeywords}
 Dynamic MR imaging, deep learning, compressed sensing, dense, local feature, global feature, total variation
\end{IEEEkeywords}

%
\IEEEpeerreviewmaketitle

\section{Introduction}
%
%
%
%
\IEEEPARstart{D}{ynamic} MR imaging is a non-invasive imaging technique which could provide both spatial and temporal information for the underlying anatomy. Nevertheless, both physiological and hardware constraints have made it suffer from slow imaging speed or long imaging time, which may lead to patients’ discomfort or sometimes cause severe motion artifacts. Therefore, it is of great necessity to accelerate MR imaging.

To accelerate dynamic MR scan, there have been three main directions of efforts, namely in developing physics based fast imaging sequences \cite{kaiser1989mr}, hardware based parallel imaging techniques \cite{sodickson1997simultaneous} and signal processing based MR image reconstruction methods from incomplete k-space data. Our specific focus here is the undersampled MR image reconstruction, which requires prior information to solve the aliasing artifacts caused by the violation of the Nyquist sampling theorem. Specifically, the reconstruction task is normally formulated as solving an optimization problem with two terms i.e. data fidelity and prior regularization. Popular prior information includes sparsity, which prompts image to be sparsely represented in a certain transform domain while being reconstructed from incoherently undersampled k-space data. These techniques are well-known as compressed sensing MRI (CS-MRI) \cite{donoho2006compressed, lustig2007sparse}. 
For example, k-t FOCUSS \cite{jung2007improved} takes advantage of the sparsity of x-f support to reconstruct x-f images from the undersampled k-t space. It encompasses the celebrated k-t BLAST and k-t SENSE \cite{tsao2003k} as special cases. And k-t ISD \cite{liang2012k} incorporates additional information on the support of the dynamic image in x-f space based on the theory of CS with partially known support. DLTG \cite{caballero2014dictionary} can learn redundancy in the data via an auxiliary constraint on temporal gradients (TG) sparsity. Wang et al \cite{wang2014compressed} employs a patch-based 3-D spatiotemporal dictionary for sparse representations of dynamic image sequence. Besides, low-rank is also a prior regularization. It can use low-rank and incoherence conditions to complete missing or corrupted entries of a matrix. A typical example on low-rank is L+S \cite{otazo2015low}, where the nuclear norm is used to enforce low rank in L and the $l_1$ norm is used to enforce sparsity in S. And k-t SLR \cite{lingala2011accelerated} exploits the correlations in the dynamic imaging dataset by modeling the data to have a compact representation in the Karhunen Louve transform (KLT) domain. These methods have made great progresses in dynamic imaging and achieved improved results. Nevertheless, these methods only draw prior knowledge from limited samples. Furthermore, the reconstruction is iterative and sometimes time-consuming. 

On the other hand, deep learning has shown great potential in accelerating MR imaging. There have been quite a few newly proposed methods, which can be roughly categorized into two types, model-based unrolling methods \cite{sun2016deep,hammernik2018learning, knoll2018assessment} and end-to-end learning methods \cite{wang2016accelerating, kwon2017parallel, han2018deep, zhu2018image, eo2018kiki,  sun2018compressed, quan2018compressed, schlemper2018deep, qin2018convolutional, wang2018image, aggarwal2018modl}. The model based unrolling methods are to formulate the iterative procedure of traditional optimization algorithms to network learning. They adaptively learn all the parameters of regularization terms and transforms in the model by network training. For example, in VN-Net \cite{hammernik2018learning}, generalized compressed sensing reconstruction formulated as a variational model is embedded in an unrolled gradient descent scheme. ADMM-Net \cite{sun2016deep} is defined over a data flow graph, which is derived from the iterative procedures in Alternating Direction Method of Multipliers (ADMM) algorithm for optimizing a CS-based MRI model. The other type utilizes the big data information to learn a network that map between the undersampled and fully sampled data pairs. Wang et al. \cite{wang2016accelerating} train a deep convolutional neural
network (CNN) to learn the mapping relationship between undersampled brain MR images and fully sampled brain MR images. AUTOMAP \cite{zhu2018image} learns a mapping between the sensor and the image domain from an appropriate training data. Despite all the successes, there are only two works that specifically apply to dynamic MR imaging \cite{schlemper2018deep, qin2018convolutional}. Both of these two works use a cascade of neural networks to learn the mapping between undersampled and fully sampled cardiac MR images, where a deep cascaded of convolutional neural network (DC-CNN) is designed in \cite{schlemper2018deep} and a convolutional recurrent neural network (CRNN) is proposed in \cite{qin2018convolutional}. Both works make great contributions to dynamic MR imaging. In our another work, we proposed a dynamic MR imaging method with both k-space and spatial prior knowledge integrated via multi-supervised network training, dubbed as DIMENSION \cite{wang2018dimension}. Although the DIMENSION model achieved the improved reconstruction results for dynamic MR imaging compared to other methods, there is still a certain degree of smooth in the reconstructed images at high acceleration factors. Part of the reasons may be the loss functions used in \cite{wang2018dimension}. The MSE loss functions only indicate the mean square information between the reconstructed image and the ground truth and cannot perceive the image structure information. Furthermore, network structures could have other options to improve the reconstruction.

In this work, we propose cascaded residual dense networks for dynamic MR imaging with edge-enhance loss constraint, dubbed as CRDN. The improvements are mainly reflected in the residual dense network structures and the total variation loss function. Our contributions could be summarized as follows:
\begin{enumerate}
	\item In this work, we propose cascaded residual dense networks for dynamic MR imaging, which can fully exploit the hierarchical features from all the convolutional layers with both local and global feature fusion. 
	\item We further utilize the total variation (TV) loss function, which has the edge enhancement properties, for training the networks. We explore the influence of different types of TV constraints on dynamic MR reconstruction, including isotropic TV, anisotropic TV and higher degree TV (HDTV). We find that the utilization of TV constraint improved the reconstruction quantitatively, among which anisotropic TV performed best.
\end{enumerate}

\section{Methodology}

\subsection{CS-MRI and CNN-MRI}
According to compressed sensing (CS) \cite{donoho2006compressed, lustig2007sparse}, MR images with a sparse representation in some transform domain can be reconstructed from randomly undersampled k-space data. Let $\textbf{S}\in \mathbb{C}^{N_xN_yN_t}$ represent a complex-valued dynamic MR image. The problem can be described by the following formula:
\begin{equation}
\label{eq_1}
\textbf{K}_u = \textbf{F}_u\textbf{S}+\textbf{e}
\end{equation}
where $\textbf{K}_u\in \mathbb{C}^{N_xN_yN_t}$ is the undersampled measurements in k-space and the unsampled points are filled with zeros. $\textbf{F}_u$ is an undersampled Fourier encoding matrix, and $\textbf{e}\in\mathbb{C}^{N_xN_yN_t}$ is the acquisition noise. We want to reconstruct $\textbf{S}$ by solving the inverse problem of Eq. \ref{eq_1}. However, the inverse problem is ill-posed, resulting in that the reconstruction is not unique. In order to reconstruct $\textbf{S}$, we constrain this inverse problem by adding some prior knowledge and solve the following optimization problem:
\begin{equation}
\label{eq_2}
\min_\textbf{S} \frac{1}{2} ||\textbf{F}_u\textbf{S}-\textbf{K}_u||_2^2+\lambda\mathcal{R}(\textbf{S})
\end{equation}
The first term is the data fidelity, which ensures that the k-space of reconstruction is consistent with the actual measurements in k-space. The second term is often referred to as the prior regularization. In the methods of CS, $\mathcal{R}(\textbf{S})$ is usually a sparse prior of $\textbf{S}$ in some transform domains, e.g. finite difference, wavelet transform and discrete cosine transformation.

In CNN-based methods, $\mathcal{R}(\textbf{S})$ is a CNN prior of $\textbf{S}$ , which force $\textbf{S}$ to match the output of the networks:
\begin{equation}
\label{eq_3}
\min_\textbf{S} \frac{1}{2} ||\textbf{F}_u\textbf{S}-\textbf{K}_u||_2^2+\lambda||\textbf{S}-f_{CNN}(\textbf{S}_u|\bm{\theta})||_2^2
\end{equation}
where $\textbf{S}_u$ is the undersampled image and $f_{CNN}(\textbf{S}_u|\bm{\theta})$ is the output of the  networks under the parameters $\bm{\theta}$. The training process of the networks is to find the optimal parameters $\bm{\theta}^*$. Once the network are trained, the networks' output $f_{CNN}(\textbf{S}_u|\bm{\theta}^*)$ is the reconstruction we want.

\subsection{The Proposed Method}

\subsubsection{The Proposed CRDN}
\begin{figure*}[!t]
	\centering
	\subfloat{\includegraphics[width=0.9\linewidth]{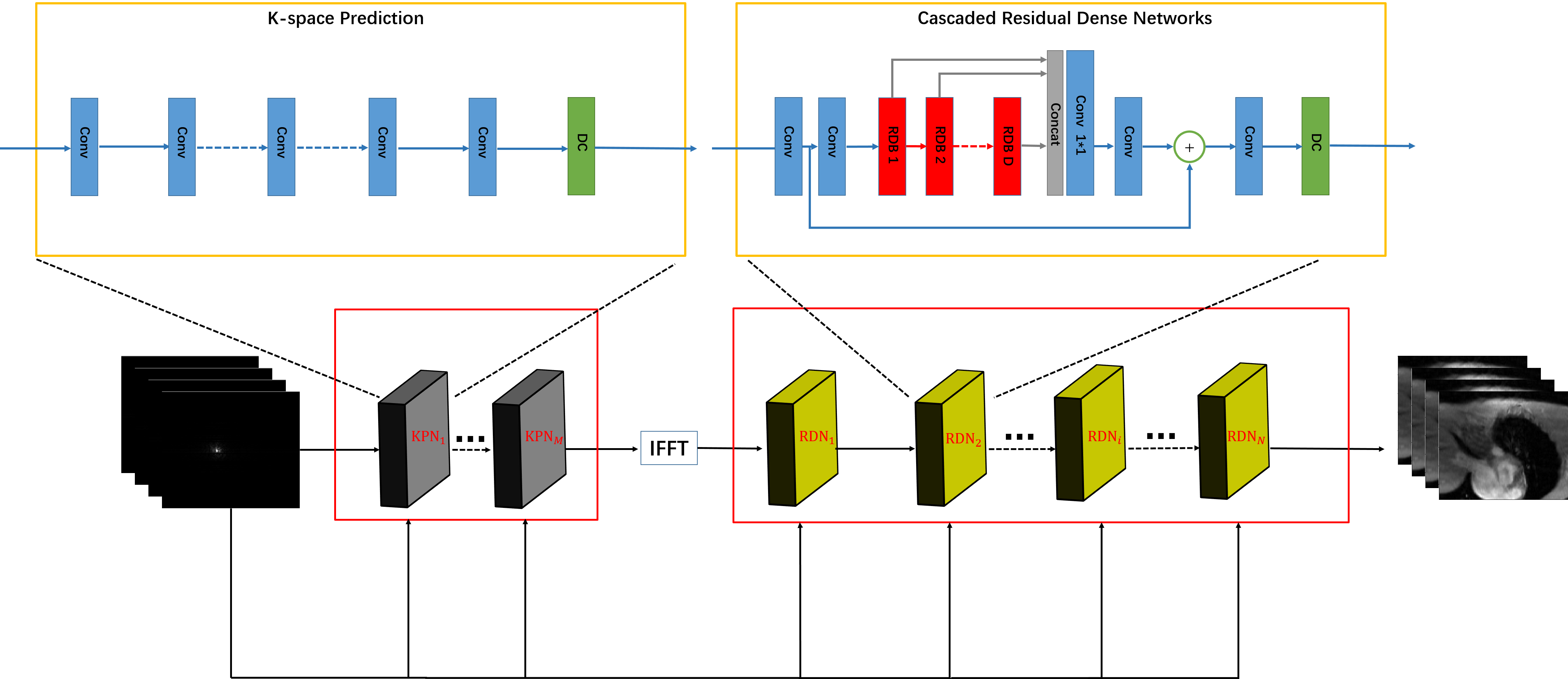}
		\label{Subnetworks}}
	
	\caption{The cascaded residual dense networks for dynamic MR imaging.}
	\label{networks}
\end{figure*}

In this work, we propose cascaded residual dense networks (CRDN) for cardiac MR image reconstruction (shown in Fig. \ref{networks}). In this work, we still adopt the idea of cross-domain learning in the DIMENSION method \cite{wang2018dimension}. The network framework includes k-space prediction networks (KPN) and cascaded residual dense networks (CRDN) in spatial domain. The two parts are connected by a Fourier inversion (see Inverse Fast Fourier Transform (IFFT) in Figure 1).
The details of the KPN can be found in \cite{wang2018dimension}. Here, we will focus on CRDN.

\begin{figure}[!t]
	\centering
	\subfloat{\includegraphics[width=1\linewidth]{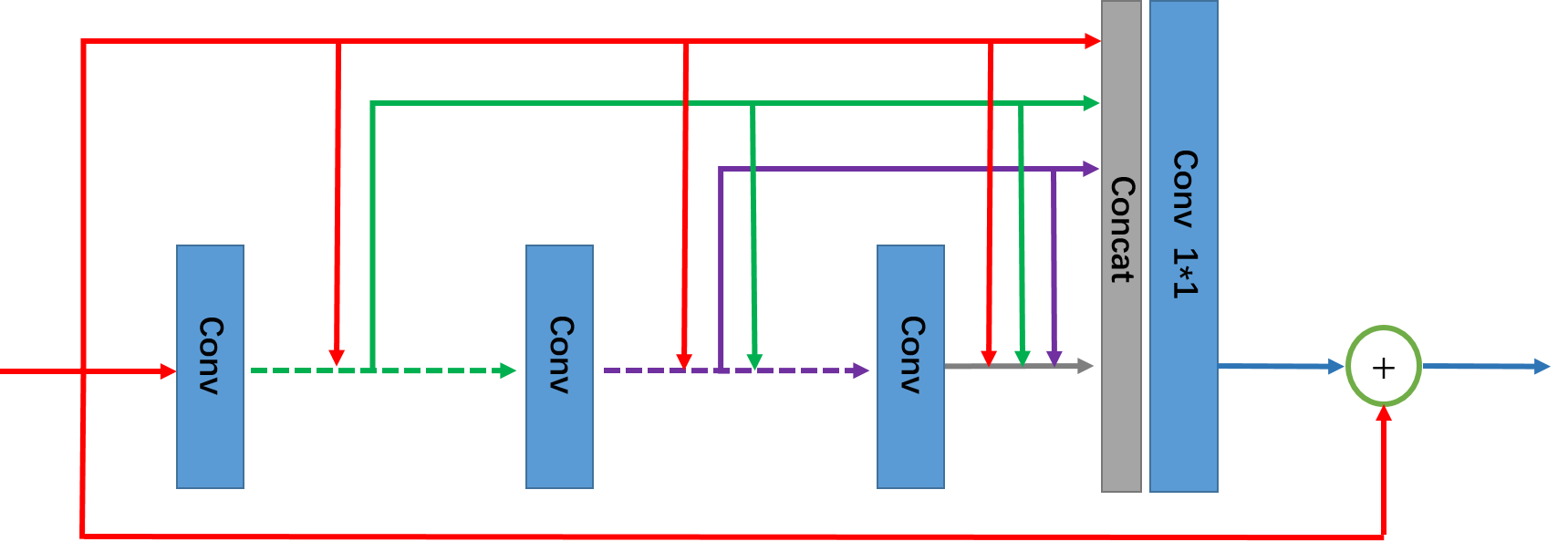}
		\label{SubRDB}}
	
	\caption{Residual dense block (RDB) for local feature fusion.}
	\label{RDB}
\end{figure}

The CRDN consists of a cascade of residual dense networks (RDN). Each RDN contains five major components: shallow feature extraction, residual dense blocks (RDBs), global feature fusion, global residual learning and data consistency (DC). Firstly, the predicted cardiac MR images by the KPN are fed into the network for shallow feature extraction. Secondly, the shallow features go through D RDBs for local feature fusion. The details of one RDB are shown in Fig. \ref{RDB}. 
The RDB includes dense connections, local feature fusion, and local residual connections. Dense connections refer to the direct connections of each convolutional layer to subsequent layers, which can enhance the transmission of local features. All local features are concatenated together and pass through a 1*1 convolutional layer to achieve local feature fusion. Residual connections are introduced in RDB to further improve information propagation. Thirdly, these residual dense features from D RDBs are merged via global feature fusion (concatenation + 1*1 convolution). We believe the combination of local feature fusion and global feature fusion can make full use of features at different levels. Fourthly, a global residual connection combines the shallow features with the global fused features. Finally, a data consistency layer \cite{schlemper2018deep} is appended to correct MR images by the accurate k-space samples.

\subsubsection{The Proposed Edge-enhanced Loss}
CNN-based methods \cite{schlemper2018deep, qin2018convolutional, wang2018dimension} can achieve improved reconstruction results in shorter time compared to classical compressed sensing (CS) or low rank based methods \cite{jung2007improved, lingala2011accelerated, otazo2015low}. However, there is still a certain degree of smooth in the reconstructed images at high acceleration factors. Part of the reasons may be the loss functions, MSE, used in these works. The MSE loss functions only indicate the mean square information between the reconstructed image and the ground truth and cannot perceive the image structure information. In this work, we explore the impacts of different total variation (TV) constraints on network training, including anisotropic TV, isotropic TV and higher degree total variation (HDTV) \cite{hu2012higher}.

Let $\hat{\textbf{S}}$ be the output of the CRDN, then the anisotropic TV can be defined as:
\begin{equation}
\label{eq_4}
TV_{aniso} (\hat{\textbf{S}})= \int_{\Omega} |\frac{\partial \hat{\textbf{S}}(r)}{\partial x}|+|\frac{\partial \hat{\textbf{S}}(r)}{\partial y}|dr
\end{equation}
And the isotropic TV can be defined as:
\begin{equation}
\label{eq_5}
TV_{iso} (\hat{\textbf{S}})= \int_{\Omega}\sqrt{(\frac{\partial \hat{\textbf{S}}(r)}{\partial x})^2+(\frac{\partial \hat{\textbf{S}}(r)}{\partial y})^2}dr
\end{equation}
The derivation and proof of HDTV can be seen in \cite{hu2012higher}. Here, we directly give out the 2 degree TV (2DTV) and 3 degree TV (3DTV):
\begin{equation}
\label{eq_6}
2DTV(\hat{\textbf{S}})= \int_{\Omega}\sqrt{(3|\hat{\textbf{S}}_{xx}|^2+3|\hat{\textbf{S}}_{yy}|^2+4|\hat{\textbf{S}}_{xy}|^2+2\mathfrak{R}(\hat{\textbf{S}}_{xx}\hat{\textbf{S}}_{yy}))/8}dr
\end{equation}
\begin{equation*}
\label{eq_7}
3DTV(\hat{\textbf{S}})= \int_{\Omega}\sqrt{5(|\hat{\textbf{S}}_{xxx}|^2+\hat{\textbf{S}}_{yyy}|^2)+6\mathfrak{R}(\hat{\textbf{S}}_{xxx}\hat{\textbf{S}}_{xyy}+\hat{\textbf{S}}_{yyy}\hat{\textbf{S}}_{xxy})+\\9(|\hat{\textbf{S}}_{xxy}|^2+\hat{\textbf{S}}_{xyy}|^2)}dr/4\sqrt{2}
\end{equation*}

\section{\textcolor{black}{EXPERIMENTAL RESULTS}}
\subsection{Setup}
\subsubsection{Data acquisition}
We collected 101 fully sampled cardiac MR data using 3T scanner (SIMENS MAGNETOM Trio) with T1-weighted FLASH sequence. Written informed consent was obtained from all human subjects. Each scan contains a single slice FLASH acquisition with 25 temporal frames. The following parameters were used for FLASH scans: FOV $330 \times 330$ mm, acquisition matrix  $192 \times 192$, slice thickness = 6 mm, TR = 3 ms, TE = 50 ms and 24 receiving coils. The raw multi-coil data of each frame was combined by adaptive coil combine method \cite{walsh2000adaptive} to produce a single-channel complex-valued image. We randomly selected 90\% of the entire dataset for training and 10\% for testing. Deep learning has a high demand for data volume \cite{lecun2015deep}. Therefore, some data augmentation strategies have been applied. We shear the original images along the $x, y$ and $t$ direction. The sheared size is $117 \times 120 \times 6 \ (x \times y \times t)$, and the stride along the three directions is 7, 7 and 5 respectively. Finally, we obtained 17500 3D complex-valued cardiac MR data with the size of $117 \times 120 \times 6$.

For each frame, the original k-space was retrospectively undersampled with 6 ACS lines. Specifically, we fully samples frequency-encodes (along $k_x$) and randomly undersamples the phase encodes (along $k_y$) according to a zero-mean Gaussian variable density function \cite{jung2007improved}. 

\subsubsection{Network training}
For network training, we divide each data into two channels, where the channels store real and imaginary parts of the data. So the inputs of the network are undersampled k-spaces $\mathbb{R}^{2N_xN_yN_t}$ and the outputs are reconstruction images $\mathbb{R}^{2N_xN_yN_t}$. In this work, we focus on a D5C5 model, which works pretty well for the DC-CNN model. The D5C5 model consists of five blocks (C5) and each block has five convolutional layers (D5). In order to simplify the parameters $\bm{\theta}$ and make a fair comparison with the D5C5 model, the KPN contains one frequency domain block and the CRDN consists of four blocks, each of which contains five convolutional layers. Therefore, both the proposed model and the D5C5 model have 25 convolutional layers in total. Therefore, both the proposed model and the D5C5 model have 25 convolutional layers in total. He initialization \cite{he2015delving} was used to initialize the network weights. Rectifier Linear Units (ReLU) \cite{glorot2011deep} were selected as the nonlinear activation functions. The mini-batch size was 20. The exponential decay learning rate \cite{zeiler2012adadelta} was used in all CNN-based experiments and the initial learning rate was set to 0.0001 with a decay of 0.95. All the models were trained by the Adam optimizer \cite{kingma2014adam} with parameters $\beta_1=0.9, \beta_2=0.999$ and $\epsilon=10^{-8}$.

The models were implemented on an Ubuntu 16.04 LTS (64-bit) operating system equipped with an Intel Xeon E5-2640 Central Processing Unit (CPU) and Tesla TITAN Xp Graphics Processing Unit (GPU, 12GB memory) in the open framework Tensorflow \cite{abadi2016tensorflow} with CUDA and CUDNN support. 

\subsubsection{Performance evaluation}
For a quantitative evaluation, mean square error (MSE), peak signal to noise ratio (PSNR) and structural similarity index (SSIM) \cite{wang2004image} were measured as follows:
\begin{equation}
\label{eq_8}
\mathrm{MSE}=
||Ref-Rec||^2_2
\end{equation}
\begin{equation}
\label{eq_9}
\mathrm{PSNR}
= 20\log_{10}\frac{\max(Ref)\sqrt{N}}{||Ref-Rec||_2}
\end{equation}
\begin{equation}
\label{eq_10}
\mathrm{SSIM}
= \boldsymbol{l}(Ref, Rec)\cdot\boldsymbol{c}(Ref, Rec)\cdot\boldsymbol{s}(Ref, Rec)
\end{equation}
where $Rec$ is the reconstructed image, $Ref$ denotes the reference image and $N$ is the total number of image pixels. The SSIM index is a multiplicative combination of the luminance term, the contrast term, and the structural term (details shown in \cite{wang2004image}).

\subsection{Does the CRDN Work?}
\begin{figure}[!t]
	\centering
	\subfloat{\includegraphics[width=1\linewidth]{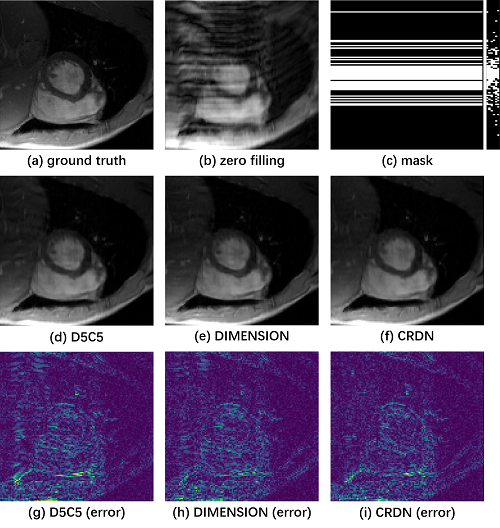}
		\label{SubKresults}}
	
	\caption{The reconstructions of the D5C5, DIMENSION and the proposed CRDN. (a) ground truth, (b) zero-filling, (c) mask and its k-t extraction, (d) the D5C5 reconstruction, (e) the DIMENSION, (f) the CRDN reconstruction, (g)-(i) their corresponding error maps with display ranges [0, 0.07].}
	\label{results}
\end{figure}

To demonstrate the efficacy of the CRDN, We compared CRDN with the D5C5 and DIMENSION methods shown in Fig. \ref{results}.

\subsection{Does the Total Variation Loss Function Work?}
To demonstrate the efficacy of the  total variation loss function, We compared different CRDN models with different types of TV shown in Fig. \ref{results_TV}.
\begin{figure*}[!t]
	\centering
	\subfloat{\includegraphics[width=1\linewidth]{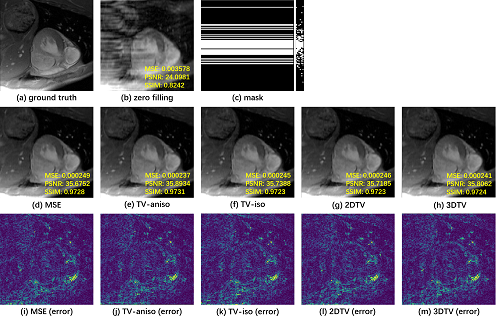}
		\label{SubKresults_TV}}
	
	\caption{The reconstructions of different types of TV loss functions}
	\label{resultsTV}
\end{figure*}

\section{Conclusion and Outlook}
  Our previous work proposed a dynamic MR imaging method with both k-space and spatial prior knowledge integrated via multi-supervised network training. Nevertheless, there was still a certain degree of smooth in the reconstructed images at high acceleration factors. In this work, we propose cascaded residual dense networks for dynamic MR imaging with edge-enhance loss constraint, dubbed as CRDN. Specifically, the cascaded residual dense networks fully exploit the hierarchical features from all the convolutional layers with both local and global feature fusion. We further utilize the total variation (TV) loss function, which has the edge enhancement properties, for training the networks. The comparisons with classical k-t FOCUSS, k-t SLR, L+S and the state-of-the-art deep learning based methods show that the improvements of network structure and loss function can achieve better reconstruction results in shorter time.

\ifCLASSOPTIONcaptionsoff
  \newpage
\fi



\bibliographystyle{IEEEtran}
\bibliography{IEEEabrv,CRDN}
\end{document}